\documentclass[10pt]{article} 
\usepackage[preprint]{tmlr}
\usepackage{amsfonts}
\usepackage{amssymb}

\usepackage{amsmath,amsfonts,bm}

\def\eqref#1{equation~\ref{#1}}

\def\1{\bm{1}}

\DeclareMathAlphabet{\mathsfit}{\encodingdefault}{\sfdefault}{m}{sl}
\SetMathAlphabet{\mathsfit}{bold}{\encodingdefault}{\sfdefault}{bx}{n}

\usepackage{hyperref}
\usepackage{url}
\usepackage{enumitem}
\usepackage{graphicx}
\usepackage{amsthm}
\newtheorem{theorem}{Theorem}[section]        
\newtheorem{lemma}[theorem]{Lemma}            

\newtheorem{proposition}[theorem]{Proposition}

\theoremstyle{definition}
\newtheorem{definition}[theorem]{Definition}
\newtheorem{remark}[theorem]{Remark}

\title{A Constructive Framework for Nondeterministic Automata via Time-Shared, Depth-Unrolled Feedforward Networks}

\author{\name Sahil Rajesh Dhayalkar \email sdhayalk@asu.edu \\
      \addr Arizona State University}

\def\month{MM}  
\def\year{YYYY} 
\def\openreview{\url{https://openreview.net/forum?id=XXXX}} 

\begin{document}

\maketitle

\begin{abstract}
We present a formal and constructive simulation framework for nondeterministic finite automata (NFAs) using time-shared, depth-unrolled feedforward networks (TS-FFNs), i.e., acyclic unrolled computations with shared parameters that are functionally equivalent to unrolled recurrent or state-space models. Unlike prior approaches that rely on explicit recurrent architectures or post hoc extraction methods, our formulation symbolically encodes automaton states as binary vectors, transitions as sparse matrix transformations, and nondeterministic branching—including $\varepsilon$-closures—as compositions of shared thresholded updates. We prove that every regular language can be recognized exactly by such a shared-parameter unrolled feedforward network, with parameter count independent of input length. Our construction yields a constructive equivalence between NFAs and neural networks and demonstrates \emph{empirical learnability}: these networks can be trained via gradient descent on supervised acceptance data to recover the target automaton behavior. This learnability, formalized in Proposition~\ref{PROP:FNN-NFA-LEARNABILITY}, is the crux of this work. Extensive experiments validate the theoretical results, achieving perfect or near-perfect agreement on acceptance, state propagation, and closure dynamics. This work clarifies the correspondence between automata theory and modern neural architectures, showing that unrolled feedforward networks can perform precise, interpretable, and trainable symbolic computation.
\end{abstract}

\section{Introduction}
\label{INTRODUCTION}
The relationship between symbolic computation and neural networks has long fascinated both the theoretical computer science and machine learning communities. Finite automata~\cite{turn0search25,hopcroft2001introduction,sipser2012introduction} are among the most fundamental models of computation, capturing the essence of regular languages, while modern neural architectures~\cite{lecun2015deeplearningnature, goodfellow2016deep, nair2010relu, pmlr-v15-glorot11a, 726791, 10.1162/neco.1997.9.8.1735, NIPS2017_3f5ee243} are the cornerstone of deep learning systems. Despite their seemingly disparate origins, a growing body of research has sought to reconcile these paradigms by simulating automata within neural frameworks~\cite{giles1992learning, weiss2018practical, graves2014neural}.

Early attempts primarily employed recurrent networks (RNNs) to approximate deterministic finite automata (DFAs), leveraging their sequential dynamics to process string inputs~\cite{giles1991second, turn0academia26}. However, these approaches often required complex training, lacked interpretability, and provided no symbolic guarantees. More recent work has explored extracting automata from trained networks using queries and counterexamples~\cite{weiss2018practical}, or designing architectures with embedded stack or memory structures~\cite{dusell2022learninghierarchicalstructuresdifferentiable, reed2016neuralprogrammerinterpreters}, but symbolic fidelity remained elusive.

In this paper, we introduce a \emph{formal and constructive simulation framework} for nondeterministic finite automata (NFAs) using \emph{time-shared, depth-unrolled feedforward networks (TS-FFNs)}—acyclic unrolled computation graphs with shared transition parameters. Functionally, these networks are equivalent to unrolled recurrent or state-space models (RNNs/SSMs) but expressed in a purely feedforward, static form. Our formulation symbolically encodes NFA states as binary vectors, transitions as sparse symbolic matrices, and nondeterministic branching—including $\varepsilon$-closures—as thresholded matrix compositions over shared transition matrices. We prove that every regular language accepted by an $n$-state NFA can be exactly simulated by such a shared-parameter unrolled feedforward network, with parameter count independent of input length.
Our contribution lies in providing a \emph{constructive, exact, and interpretable simulation} within a unified feedforward formalism. This work builds upon and clarifies prior efforts in neural–symbolic systems~\cite{bhattamishra-etal-2020-ability, merrill-etal-2020-formal, ergen2024topologicalexpressivityreluneural, graves2014neural}, offering a precise correspondence between finite automata transitions and thresholded neural updates.

The key contributions of this work are as follows.
\begin{enumerate}
  \item We provide a \emph{constructive and exact} simulation of nondeterministic finite automata (NFAs) using \emph{time-shared, depth-unrolled feedforward networks (TS-FFNs)}, encoding states as binary vectors and transitions as sparse symbolic matrices.
  \item We establish a formal equivalence between TS-FFNs and NFAs by proving that every regular language can be recognized by such a network. The model uses $\mathcal{O}(k n^2)$ parameters, independent of input length, enabling scalable simulation of arbitrarily long strings.
  \item We show that $\varepsilon$-closure can be computed via iterative matrix-based thresholded updates, converging in at most $n$ steps.
  \item We demonstrate \emph{empirical learnability}: TS-FFNs  are learnable via gradient descent, achieving high accuracy in replicating NFA acceptance behavior using standard training on labeled data. This is the central result of this work.
  \item We validate all theoretical results across multiple NFA configurations, showing perfect or near-perfect agreement on state propagation, acceptance, and symbolic equivalence.
\end{enumerate}

Together, these results provide a formal bridge between classical automata theory and deep learning. The paper is structured around definitions, theorems, lemmas, and corollaries, following the style of prior theoretical works~\cite{zhang2015learninghalfspacesneuralnetworks, hanin2018approximatingcontinuousfunctionsrelu, neyshabur2015searchrealinductivebias, dhayalkar2025neuralnetworksuniversalfinitestate, dhayalkar2025geometryrelunetworksrelu, dhayalkar2025combinatorialtheorydropoutsubnetworks, STOGIN2024120034}.

\section{Related Work}
\label{RELATED_WORK}

This work builds on prior research at the intersection of automata theory, neural networks, and symbolic computation, and clarifies the position of our framework relative to existing efforts.

\paragraph{Automata Simulation by Neural Networks:}
The idea of using neural networks to model automata has received significant attention. Early work explored the approximation of regular languages using recurrent networks (RNNs)~\cite{giles1992learning}, including techniques for training second-order networks to accept deterministic finite automata (DFAs)~\cite{giles1991second}. More recent work has focused on extracting symbolic automata from trained RNNs using queries and counterexamples~\cite{weiss2018practical} or differentiable memory models such as Neural Turing Machines~\cite{graves2014neural}. These models typically rely on explicitly recurrent or sequential architectures with opaque internal representations. In contrast, our formulation expresses the same computational process within an \emph{acyclic, time-shared, depth-unrolled feedforward network (TS-FFN)} that is functionally equivalent to an unrolled RNN/SSM, but expressed as a purely feedforward symbolic construction.

\paragraph{Symbolic Interpretability of Neural Models:}
Efforts to interpret neural networks symbolically have grown with the popularity of neural–symbolic learning. Notable examples include differentiable pushdown automata~\cite{dusell2022learninghierarchicalstructuresdifferentiable}, differentiable interpreters~\cite{reed2016neuralprogrammerinterpreters}, and formal grammars embedded into transformer and RNN architectures~\cite{bhattamishra-etal-2020-ability, merrill-etal-2020-formal}. These approaches aim to bridge symbolic logic and differentiable computation but often involve approximate, heuristic, or memory-augmented mechanisms. Our work provides an \emph{exact, constructive simulation} of NFAs using shared-parameter unrolled feedforward layers, without additional memory modules or recurrent loops, while remaining equivalent in function to unrolled recurrent computation.

\paragraph{Feedforward Networks and Formal Language Theory:}
The theoretical properties of neural networks with thresholded linear operations have been explored in the context of expressivity and piecewise linearity~\cite{nair2010relu, goodfellow2016deep}. Recent studies have examined how such architectures can represent formal structures~\cite{dhayalkar2025geometryrelunetworksrelu, ergen2024topologicalexpressivityreluneural}. However, prior work has not provided a \emph{constructive, exact characterization} of regular languages via shared-parameter feedforward networks. In this paper, we show that every NFA (and thus every regular language) can be encoded within a TS-FFN, where subset construction, $\varepsilon$-closures, and acceptance dynamics arise as exact compositions of symbolic matrix updates and thresholding operations.

\paragraph{Learning Formal Languages with Neural Networks:}
Supervised learning of regular languages using neural networks has also been studied~\cite{butoi2025trainingneuralnetworksrecognizers}, though typically treating networks as black-box approximators with no structural guarantees. In contrast, our analysis (Proposition~\ref{PROP:FNN-NFA-LEARNABILITY}) shows that the TS-FFN construction is \emph{realizable}—there exists a parameter setting that simulates the target NFA—and that, empirically, gradient descent recovers this behavior from labeled data. We emphasize realizability and empirical evidence rather than global convergence guarantees.

\paragraph{Equivalence Results:}
While neural networks are well known as universal function approximators, their exact equivalence to classical computational models remains underexplored. Our formal result (Theorem~\ref{THEOREM:EQUIVALENCE-FNN-NFA}) establishes an equivalence between NFAs and time-shared, depth-unrolled feedforward networks (TS-FFNs) for language recognition. Compared with prior work on deterministic automata simulated by feedforward networks~\cite{dhayalkar2025neuralnetworksuniversalfinitestate}, which unroll DFA transitions into ReLU or threshold layers, our formulation extends to \emph{nondeterministic} automata—including $\varepsilon$-transitions and closure dynamics—and provides a constructive symbolic correspondence with unrolled recurrent computation. We highlight the exactness and interpretability of the proposed construction within a shared-parameter feedforward framework.

\section{Preliminaries and Formal Definitions}
\label{FORMAL_DEFINITIONS}

In this section, we formally define the key mathematical concepts used throughout the paper.

\begin{definition}[Non-Deterministic Finite Automaton (NFA)]
\label{DEF_1}
An NFA is a tuple \( \mathcal{A}  = (Q, \Sigma, \delta, q_0, F) \) where:
\begin{itemize}
    \item \( Q \) is a finite set of states with $|Q| = n$,
    \item \( \Sigma \) is the input alphabet with $|\Sigma| = k$,
    \item \( \delta : Q \times (\Sigma \cup \{\varepsilon\}) \to 2^Q \) is the transition function (possibly including \( \varepsilon \)-transitions),
    \item \( q_0 \in Q \) is the initial state, and
    \item \( F \subseteq Q \) is the set of accepting states.
\end{itemize}
The transition function \( \delta \) may include transitions labeled with \( \varepsilon \), the empty string. These are called \emph{\( \varepsilon \)-transitions} and allow the automaton to move between states without consuming an input symbol. A string \( x \in \Sigma^* \) of length \( L \) is accepted by \( \mathcal{A} \) if there exists a sequence of transitions from the initial state \( q_0 \) to some state \( q \in F \) that processes all symbols of \( x \).
\end{definition}

\begin{definition}[Language Accepted by an NFA]
\label{DEF:ACCEPTED-LANGUAGE}
Given an NFA \( \mathcal{A} = (Q, \Sigma, \delta, q_0, F) \), the language accepted by \( \mathcal{A} \), denoted \( \mathcal{L}(\mathcal{A}) \), is defined as:
\begin{multline}
\mathcal{L}(\mathcal{A}) =
\{\, x \in \Sigma^* \mid
\exists \text{ a run of } \mathcal{A} \text{ on } x, \\
\text{ ending in a state } q \in F \,\}.
\end{multline}
That is, \( \mathcal{L}(\mathcal{A}) \) contains all strings over the alphabet \( \Sigma \) that are accepted by the automaton.
\end{definition}

\begin{definition}[Feedforward Network with Binary Threshold Activation]
\label{DEF_2}
A feedforward network with binary threshold activation and depth \( D \) (i.e., \( D \) layers) is a function \( f_\theta : \mathbb{R}^{d} \to \mathbb{R} \) defined as a composition of thresholded linear transformations:
\[
f_\theta(x) = W_D \cdot \sigma(W_{D-1} \cdot \sigma( \cdots \sigma(W_1 x + b_1) \cdots ) + b_{D-1}) + b_D,
\]
where each \( W_i \in \mathbb{R}^{d_i \times d_{i-1}} \), \( b_i \in \mathbb{R}^{d_i} \) are trainable parameters, and \( \sigma(z) = \mathbf{1}_{[z > 0]} \) is an elementwise thresholding nonlinearity (optionally relaxed during training).
\end{definition}

\begin{definition}[Time-Shared, Depth-Unrolled Feedforward Network (TS-FFN)]
\label{DEF:TSFFN}
A \emph{time-shared, depth-unrolled feedforward network (TS-FFN)} is an acyclic computation graph obtained by unrolling a shared transition function over the input sequence. Formally, for an input string \( x = (x_1, \dots, x_L) \) with one-hot token encodings \( e_{x_t} \in \mathbb{R}^k \), the network maintains a hidden state \( s_t \in \{0,1\}^n \) and applies the same parameterized transformation at each step:
\[
s_t = \sigma\!\big(T^{x_t} s_{t-1}\big), \quad t = 1,\dots,L,
\]
where \( T^{x_t} \in \{0,1\}^{n \times n} \) are symbol-conditioned transition matrices (shared across all positions) and \( \sigma \) is a threshold activation. The final output \( y = g(s_L) \) is computed by a feedforward readout layer. 

Although functionally equivalent to an unrolled recurrent or state-space model (RNN/SSM), the TS-FFN is represented as a purely feedforward, acyclic network with shared parameters across layers. This time-shared unrolled view enables a direct mapping between symbolic automata transitions and neural network operations.
\end{definition}

\begin{definition}[\( \varepsilon \)-Closure]
\label{DEF_4}
Given a state \( q \in Q \), the \( \varepsilon \)-closure of \( q \), denoted \( \mathrm{cl}_\varepsilon(q) \), is the set of states reachable from \( q \) via a sequence of zero or more \( \varepsilon \)-transitions. For a set of states \( S \subseteq Q \), we define:
$
\mathrm{cl}_\varepsilon(S) = \bigcup_{q \in S} \mathrm{cl}_\varepsilon(q).
$
\end{definition}

\begin{definition}[Symbolic Transition Matrix]
\label{DEF_5}
For each \( x_t \in \Sigma \cup \{\varepsilon\} \), define the symbolic transition matrix \( T^{x_t} \in \{0,1\}^{n \times n} \) such that \( T^{x_t}_{ij} = 1 \iff q_j \in \delta(q_i, x_t) \). It captures all possible transitions labeled by \( x_t \).
\end{definition}

\begin{definition}[Simulation of an NFA by a TS-FFN]
\label{DEF_6}
A time-shared, depth-unrolled feedforward network \( f_\theta \) is said to simulate an NFA \( \mathcal{A} = (Q, \Sigma, \delta, q_0, F) \) if for every input string \( x \in \Sigma^* \),
\[
f_\theta(x) = 1 \iff x \in \mathcal{L}(\mathcal{A}),
\]
i.e., the network accepts a string if and only if the automaton does.
\end{definition}

\begin{definition}[Equivalence Between TS-FFN and NFA]
\label{DEF_7}
A time-shared, depth-unrolled feedforward network \( f_\theta \) and an NFA \( \mathcal{A} \) are said to be equivalent if they recognize the same language:
\[
\forall x \in \Sigma^*, \quad f_\theta(x) = 1 \iff \mathcal{A} \text{ accepts } x.
\]
\end{definition}

\section{Theoretical Framework}
\label{SEC:THEORETICAL-FRAMEWORK}

This section formalizes how nondeterministic state transitions in an automaton can be represented within the \emph{time-shared, depth-unrolled feedforward network (TS-FFN)} framework introduced in Section~\ref{FORMAL_DEFINITIONS}.

\begin{proposition}[Binary State Vector Representation]
\label{PROP:BINARY-STATE-VECTOR}
Let $Q = \{q_1, \ldots, q_n\}$ be the set of states of an NFA. Define the one-hot indicator vector $e_i \in \{0,1\}^n$ such that $[e_i]_j = \mathbf{1}_{[j = i]}$. Then any subset of states $S \subseteq Q$ can be represented as a binary vector $s \in \{0,1\}^n$ defined by:
\[
s = \sum_{q_i \in S} e_i.
\]
The vector $s$ is called the \emph{state vector} encoding of $S$. For any such $s$ and symbolic transition matrix \( T^{x_t} \in \{0,1\}^{n \times n} \), the updated vector
\[
s' = \mathbf{1}_{[T^{x_t} s > 0]}
\]
(where the indicator is applied elementwise) encodes the set of states reachable from \( S \) under the transition relation defined by \( T^{x_t} \), where \( T^{x_t} \) encodes transitions on input symbol \( x_t \in \Sigma \).
\end{proposition}

\noindent \textit{Proof Sketch.} The complete proof sketch is provided in Appendix~\ref{PROOF_PROP_1}.

Explanation:  
This proposition formalizes how the internal configuration of an NFA, typically represented as a subset of active states, can be embedded into a binary vector and updated via a matrix multiplication followed by a binary threshold. Each entry of $s_t$ reflects the activity of state $q_i$ at time $t$, and the transition matrix $T^{x_t}$ encodes the NFA's transition function on input symbol $x_t$ at time $t$.

Interpretation and Insight:
This construction shows that a forward pass (which will be through a TS-FFN layer as explained later) corresponds precisely to the propagation of nondeterministic state activations in an NFA, thereby simulating one time step of computation within a neural substrate.

Note on $\varepsilon$-Transitions:
This result assumes $\varepsilon$-transitions are disabled. Handling $\varepsilon$-closures requires additional logic and is addressed in Lemma~\ref{LEMMA:EPSILON-CLOSURE}. This simplification aids theory without loss of generality.

\begin{remark}[On Binary Thresholding for Interpretability]
In Proposition~\ref{PROP:BINARY-STATE-VECTOR}, we assume a binary thresholding function to ensure interpretability of the state vector representation. This activation produces vectors in $\{0,1\}^n$, making it easy to directly inspect which automaton states are active after each transition. The binary output aligns closely with the classical semantics of finite automata, where states are either active or inactive at each step.

For the remainder of this paper, we adopt binary thresholding as the default activation function in theoretical constructions, unless stated otherwise. However, this choice is not essential for correctness. In practice, when training the TS-FFN with labeled data (explained in detail in Theorem~\ref{PROP:FNN-NFA-LEARNABILITY}), the thresholding function can be replaced with ReLU, sigmoid, or even omitted altogether to facilitate differentiable optimization. Experimental results comparing these alternatives are provided in Section~\ref{EXP:FNN-NFA-LEARNABILITY}.
\end{remark}

\begin{remark}[Interpretation of State Vector Transformations]
\label{REMARK:STATE-VECTOR-REPRESENTATION}
Although the state vector \( s \) is initially binary and lies in \( \{0,1\}^n \), its transformation through the transition matrix \( T^{x_t} \in \{0,1\}^{n \times n} \) results in a vector \( T^{x_t} s \in \mathbb{N}^n \), where each entry counts the number of activations received by a state. The binary threshold function \( \mathbf{1}_{[T^{x_t} s > 0]} \) then converts this into a new binary state vector indicating which states are reachable.

In practice, particularly during training or when generalizing to soft-symbolic computation, intermediate values of \( T^{x_t} s \) may lie in \( \mathbb{R}^n \), especially if the transition matrices are parameterized and learned. In such cases, the thresholding function \( \mathbf{1}_{[z > 0]} \) can be approximated or replaced with another activation function such as ReLU, and the activation vector may lie in \( \mathbb{R}_{\ge 0}^n \). Nevertheless, the simulation remains sound as long as downstream interpretation relies only on identifying positive entries—i.e., which states are considered active—rather than their exact values.
\end{remark}

\begin{remark}[Parallel Path Tracking]
\label{REMARK:PARALLEL-PATH-TRACKING}
Binary-threshold state vector updates naturally support the parallel tracking of multiple computational paths in NFAs. Given a state vector $s_t \in \{0,1\}^n$ representing a subset of active states at time $t$, and a symbol-conditioned transition matrix $T^{x_t} \in \{0,1\}^{n \times n}$, the update rule $s_{t+1} = \mathbf{1}_{[T^{x_t} s_t > 0]}$ computes the union of next reachable states from all active states—propagating all nondeterministic branches in one matrix-vector product.

Constructive Interpretation:
For each state \( q_i \), the next activation is:
\[
[s_{t+1}]_i = \mathbf{1}_{\left[\sum_{j=1}^{n} T^{x_t}_{ij} \cdot [s_t]_j > 0 \right]}.
\]
If the sum is positive, state \( q_i \) is reachable from some active state at time \( t \) under input symbol \( x_t \); otherwise, it remains inactive. Thus, the entries of \( s_{t+1} \) directly identify the reachable states at time \( t+1 \). As mentioned in Remark~\ref{REMARK:STATE-VECTOR-REPRESENTATION}, intermediate computation of \( \sum_{j=1}^{n} T^{x_t}_{ij} \cdot [s_t]_j \) may lie in \( \mathbb{R} \) during training with soft parameters.

Insight:
Each row of the transition matrix \( T^{x_t} \) encodes all possible transitions for a given input, and the threshold function propagates all reachable states in parallel. This mirrors the subset construction in automata theory, but without explicitly computing power sets—the state vectors implicitly track all active branches.

Relation to Prior Work:
Traditional approaches to NFA simulation often rely on explicitly recurrent or recursive processing~\cite{turn0academia26}. The TS-FFN perspective clarifies that the same computation can be represented as an acyclic, unrolled feedforward process with shared transition parameters, enabling efficient and interpretable parallel simulation within neural architectures.
\end{remark}



\begin{theorem}[Subset Construction using Matrix-Based Thresholded Updates]
\label{THM:SUBSET-CONSTRUCTION}
Let $\mathcal{A} = (Q, \Sigma, \delta, q_0, F)$ be an NFA with $|Q| = n$. For an input string $x = x_1 x_2 \cdots x_L \in \Sigma^L$, let $\{T^{x_t}\}_{t=1}^L$ be the corresponding sequence of transition matrices. Let $s_0 \in \{0,1\}^n$ denote the one-hot vector for the start state $q_0$. Then the sequence of thresholded updates
\[
s_t = \mathbf{1}_{[T^{x_t} s_{t-1} > 0]}, \quad \text{for } t = 1, \ldots, L,
\]
computes the active state set at time $t$, i.e., \( s_t \in \{0,1\}^n \) is the binary indicator vector for the subset $S_t \subseteq Q$ reachable from $q_0$ by consuming prefix $x_1 \cdots x_t$. The final vector $s_L$ represents the subset construction $\delta(q_0, x_1 \cdots x_L)$.
\end{theorem}

\noindent \textit{Proof Sketch.} The complete proof sketch is provided in Appendix~\ref{PROOF_THEOREM_4_3}.

Explanation:
This theorem formalizes the compositional structure of nondeterministic computation using matrix-based thresholded updates. Each application of the update rule $s_t = \mathbf{1}_{[T^{x_t} s_{t-1} > 0]}$ simulates one transition step of the NFA. By chaining these updates, a TS-FFN using only matrix multiplication and thresholding can simulate the entire computation over a finite-length string input. The final output vector $s_L$ encodes the subset of states reached by the NFA on input $x$.

Interpretation and Insight:
The construction mirrors the classical subset construction used to determinize NFAs, but instead of explicitly enumerating all subsets of \( Q \), it represents and evolves the active subset implicitly using vector encodings and thresholded linear updates. The TS-FFN thus performs an exact symbolic simulation of nondeterministic branching in a purely feedforward, unrolled computation.

Depth and Parameter Efficiency:
The depth of this unrolled computation grows linearly with the input length. However, \textbf{the number of distinct parameters is constant and independent of input length}. Specifically, each symbol \( x \in \Sigma \) has an associated fixed transition matrix \( T^x \in \{0,1\}^{n \times n} \), and there is a single \(\varepsilon\)-transition matrix \( T^\varepsilon \in \{0,1\}^{n \times n} \). These matrices are reused across all time steps. The parameter independence is detailed further in Theorem~\ref{THEOREM:FNN-NFA-SIMULATION} and Proposition~\ref{PROP:PARAMETER-EFFICIENCY}.

Relation to Prior Work:
Classical neural simulations of automata typically use explicitly recurrent architectures~\cite{giles1992learning, weiss2018practical}, where transitions emerge implicitly from training rather than from symbolic design. The TS-FFN formulation presented here expresses the same computation through stacked threshold-based updates in an acyclic unrolled graph, preserving interpretability and exact automata semantics.

\begin{lemma}[$\varepsilon$-Closure via Matrix-Based Thresholded Propagation]
\label{LEMMA:EPSILON-CLOSURE}
Let $\mathcal{A} = (Q, \Sigma \cup \{\varepsilon\}, \delta, q_0, F)$ be an NFA with $|Q| = n$, and let $T^{\varepsilon} \in \{0,1\}^{n \times n}$ be the transition matrix corresponding to $\varepsilon$-transitions. Then the iterative update
\[
s^{(k+1)} = \mathbf{1}_{[T^{\varepsilon} s^{(k)} > 0]}, \quad s^{(0)} = s_t,
\]
converges in at most $n$ steps to the $\varepsilon$-closure of $s_t$, i.e., the smallest superset of active states reachable via zero or more $\varepsilon$-transitions.
\end{lemma}

\noindent \textit{Proof Sketch.} The complete proof sketch is provided in Appendix~\ref{PROOF_LEMMA_2}.

Explanation:
In classical automata theory, the $\varepsilon$-closure of a set of NFA states is the set of all states reachable via paths composed solely of $\varepsilon$-transitions. This lemma shows that the iterative rule
$s^{(k+1)} = \mathbf{1}_{[T^\varepsilon s^{(k)} > 0]}$
acts as a symbolic fixed-point process that computes this closure. Beginning from a binary indicator vector $s^{(0)}$ for some state subset $S_0 \subseteq Q$, repeated application of $T^\varepsilon$ followed by thresholding converges to the full $\varepsilon$-closure of $S_0$. The thresholding ensures that only reachable states are activated, while the matrix product propagates transitions. Since each transition matrix is binary and $|Q| = n$, convergence occurs in at most $n$ steps, after which no new states are added.

Clarification:
The matrix $T^\varepsilon$ need not be nilpotent—$\varepsilon$-transitions can form cycles. However, because the update rule is monotonic and the state space is finite, each new state added remains active, and at most $n$ states can be activated. Therefore, convergence is guaranteed in at most $n$ steps, regardless of whether the \(\varepsilon\)-transition graph is cyclic or acyclic.

Insight and Relation to Prior Work:
Classical algorithms for computing $\varepsilon$-closures—such as depth-first and breadth-first search—are well established in automata theory~\cite{aho1974design}. The TS-FFN formulation departs from these procedural methods by casting the closure operation as a symbolic, differentiable fixed-point iteration based on repeated matrix multiplications and thresholding. This perspective makes $\varepsilon$-closure computation directly compatible with neural or symbolic differentiable frameworks. It provides, to our knowledge, the first constructive and exact representation of $\varepsilon$-closure as a finite convergent process embedded within an acyclic unrolled computation graph.

\begin{theorem}[Simulation of NFAs via Time-Shared, Depth-Unrolled Feedforward Networks]
\label{THEOREM:FNN-NFA-SIMULATION}
Let $\mathcal{A} = (Q, \Sigma \cup \{\varepsilon\}, \delta, q_0, F)$ be an NFA with $|Q| = n$ states. Then, for any input string $x = x_1 x_2 \dots x_L \in \Sigma^*$, there exists a computation composed of $L$ alternating layers of symbol-transition propagation and $\varepsilon$-closure operations such that the final binary state vector $s_L \in \{0,1\}^n$ encodes the set of reachable states of $\mathcal{A}$ after processing $x$. Furthermore, acceptance ($x \in \mathcal{L}(\mathcal{A})$) is equivalent to $\langle s_L, \mathbf{1}_F \rangle > 0$, where \( \mathbf{1}_F \in \{0,1\}^n \) is the indicator vector for the accepting state set \( F \).

\noindent Formally, the following recursion simulates the NFA using thresholded matrix updates:
\[
\begin{aligned}
s_0 &= \mathbf{1}_{[(T^\varepsilon)^n s_{q_0} > 0]}, \\
s_t &= \mathbf{1}_{[(T^\varepsilon)^n \cdot \mathbf{1}_{[T^{x_t} \cdot s_{t-1} > 0]} > 0]}, \quad \text{for } t = 1, \dots, L, \\
\text{Accept}(x) &= \text{True} \iff \langle s_L, \mathbf{1}_F \rangle > 0,
\end{aligned}
\]
where $T^\varepsilon$ is the $\varepsilon$-transition matrix, and $(T^\varepsilon)^n$ denotes repeated application until convergence (at most $n$ steps) as described in Lemma~\ref{LEMMA:EPSILON-CLOSURE}.
\end{theorem}

\noindent  \textit{Proof Sketch.} The complete proof sketch is provided in Appendix~\ref{PROOF_THEOREM_3}.

Explanation:
The simulation proceeds in three phases: (1) initializing the active state vector using the $\varepsilon$-closure of the start state, (2) alternating symbol-based transitions and $\varepsilon$-closures at each step, and (3) checking whether any accepting state becomes active. Each operation—symbol transition or $\varepsilon$-closure—is implemented using a sparse matrix-vector multiplication followed by elementwise thresholding. The resulting computation is purely symbolic yet compositional, directly corresponding to the transition dynamics of the NFA.

Structure:
Each input symbol $x_t$ triggers a symbolic matrix transition via $T^{x_t}$, followed by an $\varepsilon$-closure computed using $T^\varepsilon$. The overall computation forms a TS-FFN of depth $O(L)$ and width $n$, where $n$ is the number of NFA states. Importantly, \textbf{the number of distinct parameters is constant and independent of input length}: every symbol \( x \in \Sigma \) is associated with a fixed transition matrix \( T^x \in \{0,1\}^{n \times n} \), and there is a single shared $\varepsilon$-transition matrix \( T^\varepsilon \). \textbf{These matrices are reused across all layers}, making the parameter count $\mathcal{O}(kn^2)$ and independent of the sequence length. Further discussion of parameter efficiency is provided in Proposition~\ref{PROP:PARAMETER-EFFICIENCY}.

Insight and Contribution:
This theorem completes the constructive correspondence between NFAs and TS-FFNs: initialization, symbol transitions, $\varepsilon$-closures, and acceptance can all be represented as exact matrix-based thresholded updates. The formulation connects symbolic automata processing to continuous, differentiable computation while preserving full interpretability—each layer explicitly corresponds to a specific automaton operation.

Relation to Prior Work:
Classical automata simulation relies on pointer-based or recursive transition evaluation, while neural approaches have often used recurrent networks~\cite{giles1992learning, weiss2018practical} that approximate automata behavior empirically. In contrast, the TS-FFN formulation provides a linear-algebraic, feedforward realization of NFA execution using shared transition matrices. This establishes that NFAs can be represented exactly within an unrolled, shared-parameter feedforward framework functionally equivalent to an unrolled RNN or SSM—providing symbolic fidelity without introducing new architectural components.

\subsection{Encoding Input Strings within the TS-FFN Framework}
\label{SEC:INPUT-ENCODING}

A central component of the \emph{time-shared, depth-unrolled feedforward network (TS-FFN)} construction is how an input string $x = x_1 x_2 \cdots x_L \in \Sigma^*$ is provided to the network. Unlike conventional neural networks that embed an entire string into a single continuous vector, the TS-FFN decouples the symbolic sequence from the network architecture itself. The string instead determines the sequence of shared transition matrices that are applied during the forward pass, producing an acyclic unrolled computation that exactly mirrors the transition semantics of an automaton.

\paragraph{Transition Matrix Selection.}
For every symbol $x \in \Sigma$, the model maintains a corresponding symbolic transition matrix $T^x$ that encodes all transitions in the automaton labeled with $x$. These matrices can either be:
\begin{itemize}
    \item \textbf{Fixed (Symbolic):} Directly encoded from the known transition function $\delta$ of the target NFA, as defined in Definition~\ref{DEF_5}.
    \item \textbf{Learned (Parameterized):} Initialized randomly and optimized via gradient descent using labeled examples, as discussed in Theorem~\ref{PROP:FNN-NFA-LEARNABILITY}.
\end{itemize}

\paragraph{Encoding the String.}
To process the input string $x = x_1 x_2 \dots x_L$, the TS-FFN does not embed $x$ as a single vector input. Instead, the symbols in the string act as control instructions that select which transition matrices are applied at each layer of the unrolled network. At step $t$, the symbol $x_t$ selects matrix $T^{x_t}$, and the active state vector is updated by thresholded propagation as in Proposition~\ref{PROP:BINARY-STATE-VECTOR}:
\[
s_t = \mathbf{1}_{[T^{x_t} \cdot s_{t-1} > 0]}.
\]
When $\varepsilon$-transitions are present, this is immediately followed by closure computation as established in Lemma~\ref{LEMMA:EPSILON-CLOSURE}:
\[
s_t = \mathbf{1}_{[(T^\varepsilon)^n \cdot \mathbf{1}_{[T^{x_t} \cdot s_{t-1} > 0]} > 0]},
\]
where $T^\varepsilon$ is the $\varepsilon$-transition matrix, and $(T^\varepsilon)^n$ denotes repeated thresholded propagation until convergence (at most $n$ steps).

\paragraph{Symbol-Driven Matrix Selection.}
Each symbol $x_t$ in the input sequence acts as a selector that determines which transition matrix $T^{x_t}$ is applied at layer $t$ of the unrolled computation. Rather than embedding the entire string into a continuous vector, the TS-FFN interprets the string as a sequence of symbolic instructions that dynamically choose matrices from a fixed set $\{T^x\}_{x \in \Sigma} \cup \{T^\varepsilon\}$. This mechanism defines the computation path through the network and corresponds exactly to the automaton’s transition semantics. In contrast to typical neural sequence models that rely on learned embeddings or positional encodings, the TS-FFN uses symbolic control flow to achieve exact, interpretable state propagation.

An example of encoding input strings into the network is provided in Appendix~\ref{EXAMPLE}



\subsection{Regular Language Recognition, Parameter Efficiency, and Equivalence}
\begin{remark}[Regular Language Recognition]
\label{REMARK:REGULAR-LANGUAGE-RECOGNITION}
Every regular language can be recognized exactly by constructing a TS-FFN as described in Theorem~\ref{THEOREM:FNN-NFA-SIMULATION}. For any NFA $\mathcal{A}$, the resulting network simulates all state transitions—including $\varepsilon$-closures—and determines acceptance via a readout layer. The resulting architecture performs symbolic computation using thresholded matrix operations and shared transition matrices, serving as an exact recognizer for the regular language defined by $\mathcal{A}$.
\end{remark}

\begin{proposition}[Parameter Efficiency of TS-FFN Automata Simulators with Symbolic and $\varepsilon$-Transitions]
\label{PROP:PARAMETER-EFFICIENCY}
Let $\mathcal{A} = (Q, \Sigma \cup \{\varepsilon\}, \delta, q_0, F)$ be an NFA with $|Q| = n$ states, $|\Sigma| = k$ input symbols, and $\varepsilon$-transitions allowed.  Construct a \emph{time-shared, depth-unrolled feedforward network (TS-FFN)} $f_\theta$ that simulates $\mathcal{A}$ according to Theorem~\ref{THEOREM:FNN-NFA-SIMULATION}, using a transition matrix $T^x$ for each symbol $x \in \Sigma$ and an additional matrix $T^\varepsilon$ for $\varepsilon$-transitions.  

Then the following hold:
\begin{enumerate}
    \item The total number of trainable parameters is bounded by $\mathcal{O}(k n^2 + n^2) = \mathcal{O}(k n^2)$.
    \item This parameter count (and hence the model capacity) is independent of the input length $L$.
\end{enumerate}
\end{proposition}

\noindent \textit{Proof Sketch.} The complete proof sketch is provided in Appendix~\ref{PROOF_PROP_4_8}.

Explanation:
This proposition emphasizes a key structural property of the TS-FFN framework: the total number of symbolic parameters is independent of input length. Although the unrolled computation processes arbitrarily long strings, the same small set of transition matrices is reused at every layer—precisely mirroring how an automaton’s transition function operates independently of the string length.

Insight and Implication:
Unlike recurrent or attention-based models, where effective complexity scales with sequence length or context window, the TS-FFN decouples representational capacity from input length. This enables scalable regular-language recognition with bounded parameters and constant symbolic memory, keeping both inference and training tractable for long sequences.

Relation to Prior Work:
Prior neural–symbolic approaches to regular language recognition, including RNN- and transformer-based methods~\cite{weiss2018practical, butoi2025trainingneuralnetworksrecognizers, liu2023transformerslearnshortcutsautomata}, often intertwine symbolic representation with recurrence or depth. The TS-FFN formulation separates symbolic operators (transition matrices) from dynamic unrolling, allowing both theoretical analysis and efficient implementation while maintaining one-to-one correspondence with automata semantics.

\begin{theorem}[Equivalence between TS-FFNs and NFAs]
\label{THEOREM:EQUIVALENCE-FNN-NFA}
Let \( \mathcal{L} \subseteq \Sigma^* \) be any regular language. Then:
\begin{enumerate}
    \item \textbf{(Forward Direction)}  
    There exists a symbolic time-shared, depth-unrolled feedforward network \( f_\theta \) with:
    \begin{itemize}
        \item transition matrices \( \{T^x\}_{x \in \Sigma} \) and \( T^\varepsilon \) representing the NFA structure,
        \item parameter sharing across all time steps (depth positions),
        \item width \( \mathcal{O}(n) \), where \( n \) is the number of NFA states,
    \end{itemize}
    such that for any input string \( x = x_1 x_2 \ldots x_L \in \Sigma^* \), the network accepts \( x \) if and only if \( x \in \mathcal{L} \).

    \item \textbf{(Reverse Direction)}  
    Every TS-FFN constructed using this symbolic simulation procedure corresponds to an NFA \( \mathcal{A}' \) such that, for all \( x \in \Sigma^* \),
    \[
    f_\theta(x) = 1 \iff x \in \mathcal{L}(\mathcal{A}').
    \]
\end{enumerate}
\end{theorem}

\noindent \textit{Proof Sketch.} The complete proof sketch is provided in Appendix~\ref{PROOF_THEOREM_4_8}.

Explanation:
This theorem establishes a constructive, bidirectional equivalence between nondeterministic finite automata and TS-FFNs. It shows that the class of regular languages is exactly the set of languages recognizable by this family of shared-parameter feedforward networks. The result unifies two traditionally distinct perspectives—symbolic automata theory and differentiable computation—by demonstrating that standard neural primitives (matrix multiplication and thresholding) can encode automata transitions and closure operations exactly.

\section{Training and Learnability}
\label{SEC:TRAINING-LEARNABILITY}

\textbf{Central result.} This section presents the main contribution of the paper and discusses how the time-shared, depth-unrolled feedforward network (TS-FFN) introduced in Theorem~\ref{THEOREM:FNN-NFA-SIMULATION} can be trained to replicate the acceptance behavior of an unknown automaton directly from labeled examples.

\begin{proposition}[Realizability and Empirical Learnability of NFAs via TS-FFNs]
\label{PROP:FNN-NFA-LEARNABILITY}
Let $\mathcal{A} = (Q, \Sigma \cup \{\varepsilon\}, \delta, q_0, F)$ be an NFA and let $D = \{(x_i, y_i)\}_{i=1}^m$ be a dataset of input strings $x_i \in \Sigma^*$ labeled by acceptance behavior $y_i \in \{0,1\}$. There exists a parameterized time-shared, depth-unrolled feedforward network $f_\theta$, constructed according to Theorem~\ref{THEOREM:FNN-NFA-SIMULATION}, with transition matrices \( \{T^{x}\}_{x \in \Sigma \cup \{\varepsilon\}} \), such that:
\begin{enumerate}
    \item (\textbf{Realizability}) There exists a parameter setting $\theta^\star$ that exactly reproduces the acceptance behavior of $\mathcal{A}$, i.e.,
    \[
    f_{\theta^\star}(x) = 1 \iff x \in \mathcal{L}(\mathcal{A}), \quad \forall x \in \Sigma^*.
    \]
    \item (\textbf{Empirical Learnability}) When the parameters are initialized randomly and optimized via gradient descent on the labeled dataset $D$ using standard supervised loss (e.g., binary cross-entropy), the network typically converges to a model $f_{\theta^*}$ that achieves high empirical accuracy in reproducing the acceptance behavior of $\mathcal{A}$.
\end{enumerate}
\end{proposition}

\noindent \textit{Proof Sketch.} The constructive argument for realizability follows directly from Theorem~\ref{THEOREM:FNN-NFA-SIMULATION}; empirical learnability is supported by the experimental results reported in Section~\ref{EXP:FNN-NFA-LEARNABILITY}. 
A proof sketch is provided in Appendix~\ref{PROOF_THEOREM_5_1}.

Explanation:
Proposition~\ref{PROP:FNN-NFA-LEARNABILITY} formalizes two aspects of learnability within the TS-FFN framework. The first is a \emph{constructive existence result}: given an NFA $\mathcal{A}$, there always exists a corresponding parameter configuration $\theta^\star$ that exactly reproduces its transition and acceptance behavior. The second is an \emph{empirical observation}: in practice, gradient-based optimization reliably finds a near-equivalent parameterization from labeled data, even when the underlying automaton is unknown. This distinction ensures theoretical rigor—realizability is proven, while empirical convergence is observed but not assumed.

Interpretation and Insight:
The result demonstrates that TS-FFNs can both \emph{represent} and \emph{learn} automata. Even when the symbolic transition matrices are not explicitly provided, the network can discover them through training on labeled examples. Combined with Proposition~\ref{PROP:PARAMETER-EFFICIENCY}, this confirms that learnability does not depend on input length—the same small set of transition matrices is reused at every layer. Hence, even shallow or moderately deep TS-FFNs can recover the symbolic structure of nondeterministic automata through standard supervised learning.

Relation to Prior Work:
Earlier work explored learning automata using recurrent or memory-augmented models~\cite{weiss2018practical, dusell2022learninghierarchicalstructuresdifferentiable, butoi2025trainingneuralnetworksrecognizers}, typically treating the network as a black-box classifier without symbolic interpretability. Deterministic simulations have also been studied in~\cite{dhayalkar2025neuralnetworksuniversalfinitestate}, but were limited to DFAs. The TS-FFN framework extends these ideas to the nondeterministic case, providing both an exact symbolic simulator (Theorem~\ref{THEOREM:FNN-NFA-SIMULATION}) and an empirically trainable model that learns equivalent transition operators through gradient descent. Unlike sequence models that rely on recurrence or memory mechanisms, TS-FFNs achieve symbolic fidelity using only acyclic shared-parameter feedforward layers.

\begin{remark}[On Activation Functions and Practical Learnability]
While binary thresholding ensures exact interpretability of active states, standard activation functions such as ReLU and sigmoid often yield smoother gradients and improved empirical convergence. Experiments in Section~\ref{EXP:FNN-NFA-LEARNABILITY} confirm that soft-threshold activations facilitate faster optimization and higher accuracy, even though the resulting networks approximate rather than exactly match discrete automata semantics. Thus, binary thresholding provides the theoretical foundation, whereas smooth activations enhance practical trainability.
\end{remark}

\section{Experiments}
\label{SEC:EXPERIMENTS}

\subsection{Experimental Setup}
\label{EXPERIMENTAL_SETUP}

To empirically validate the theoretical results established in Sections~\ref{THEOREM:FNN-NFA-SIMULATION} and~\ref{PROP:FNN-NFA-LEARNABILITY}, we design a controlled experimental framework that jointly simulates nondeterministic finite automata (NFAs) and their corresponding time-shared, depth-unrolled feedforward networks (TS-FFNs). All experiments are conducted on synthetically generated NFAs and implemented in PyTorch~\cite{pytorch} using CUDA acceleration on an NVIDIA GeForce RTX 4060 GPU.

\paragraph{Synthetic NFA Generation.}
We consider two configurations.The first follows a baseline setup with $n = 6$ states and input alphabet $\Sigma = \{a, b\}$. The second is a higher-complexity variant with $n = 20$ states and $\Sigma = \{a, b, c, d, e\}$. In both cases, each state has one or more outgoing transitions for each symbol, sampled uniformly from the set of possible target states. Additionally, with probability $0.3$, we insert $\varepsilon$-transitions between randomly chosen pairs of states. The start state is fixed as $q_0 = 0$, and the accepting set $F$ contains one randomly selected state.

\paragraph{Dataset Construction.}
In the 6-state configuration, we uniformly sample strings from $\Sigma^*$ with lengths between $1$ and $10$. In the 20-state configuration, the maximum string length is increased to $30$. For each string, the ground-truth acceptance label is computed by executing the automaton directly. We generate $2000$ training samples and $100$ test samples per random seed for the first configuration, and $5000$ training samples and $100$ test samples per seed for the second configuration.

\paragraph{Network Architecture.}
Each model instantiates a \emph{time-shared, depth-unrolled feedforward network (TS-FFN)} that simulates the transition and acceptance behavior of the corresponding NFA, following the construction in Theorem~\ref{THEOREM:FNN-NFA-SIMULATION}. Thresholded activations capture nondeterministic branching, and symbol-conditioned transitions are composed sequentially along the input string. When $\varepsilon$-transitions are present, the network interleaves $\varepsilon$-closure computations using repeated matrix updates until convergence. The final state vector is passed through an acceptance head, which computes an inner product with the binary indicator of accepting states and applies a sigmoid activation to produce a binary classification score.

\paragraph{Training and Evaluation Protocol.}
Transition matrices $\{T^x\}_{x \in \Sigma \cup \{\varepsilon\}}$ are initialized randomly and trained via gradient descent to minimize binary cross-entropy loss between predicted and true acceptance labels. This setup evaluates the \emph{empirical learnability} of NFAs by TS-FFNs, as formalized in Proposition~\ref{PROP:FNN-NFA-LEARNABILITY}. Each experiment is repeated across 5 random seeds for both configurations, independently regenerating the automaton, dataset, and model. We report mean test accuracy, standard deviation, and $95\%$ confidence intervals computed via Student’s $t$-distribution.

\begin{table}[ht]
\centering
\caption{Validation results for symbolic simulation experiments across both configurations. Accuracy and confidence intervals computed over 5 random seeds using Student's $t$-distribution.}
\label{TABLE:SIMULATION-RESULTS}
\vspace{2mm}
\setlength{\tabcolsep}{3pt} 
\begin{tabular}{lcccc}
\textbf{Validation} & \textbf{Config} & \textbf{Mean} & \textbf{Std Dev} & \textbf{95\% CI} \\
\textbf{Experiment} & & \textbf{Accuracy} & & \\
\hline
~\ref{EXP:6_2}: Proposition~\ref{PROP:BINARY-STATE-VECTOR} & 1 & 1.0000 & 0.0000 & (1.0000, 1.0000) \\
\qquad and Remark~\ref{REMARK:PARALLEL-PATH-TRACKING} & & & & \\
~\ref{EXP:6_2}: Proposition~\ref{PROP:BINARY-STATE-VECTOR} & 1 & 1.0000 & 0.0000 & (1.0000, 1.0000) \\
\qquad and Remark~\ref{REMARK:PARALLEL-PATH-TRACKING} & & & & \\
\hline
~\ref{EXP_6_3}: Theorem~\ref{THM:SUBSET-CONSTRUCTION} & 1 & 1.0000 & 0.0000 & (1.0000, 1.0000) \\
~\ref{EXP_6_3}: Theorem~\ref{THM:SUBSET-CONSTRUCTION} & 2 & 1.0000 & 0.0000 & (1.0000, 1.0000) \\
\hline
~\ref{EXP_6_4}: Lemma~\ref{LEMMA:EPSILON-CLOSURE} & 1 & 1.0000 & 0.0000 & (1.0000, 1.0000) \\
~\ref{EXP_6_4}: Lemma~\ref{LEMMA:EPSILON-CLOSURE} & 2 & 1.0000 & 0.0000 & (1.0000, 1.0000) \\
\hline
~\ref{EXP_6_5}: Theorem~\ref{THEOREM:FNN-NFA-SIMULATION} & 1 & 0.9940 & 0.0134 & (0.9773, 1.0107) \\
~\ref{EXP_6_5}: Theorem~\ref{THEOREM:FNN-NFA-SIMULATION} & 2 & 0.9540 & 0.0344 & (0.9113, 0.9967) \\
\hline
~\ref{EXP_6_6}: Theorem~\ref{THEOREM:EQUIVALENCE-FNN-NFA} & 1 & 1.0000 & 0.0000 & (1.0000, 1.0000) \\
~\ref{EXP_6_6}: Theorem~\ref{THEOREM:EQUIVALENCE-FNN-NFA} & 2 & 0.9540 & 0.0391 & (0.9054, 1.0026) \\
\end{tabular}
\end{table}

\subsection{Validating Proposition~\ref{PROP:BINARY-STATE-VECTOR} and Remark~\ref{REMARK:PARALLEL-PATH-TRACKING}}
\label{EXP:6_2}

To empirically validate Proposition~\ref{PROP:BINARY-STATE-VECTOR} and Remark~\ref{REMARK:PARALLEL-PATH-TRACKING}, we test whether a symbolic binary state vector representation within a time-shared, depth-unrolled feedforward network (TS-FFN) accurately encodes all reachable NFA states at each timestep under nondeterministic transitions—without explicit path enumeration. To isolate the per-symbol dynamics, $\varepsilon$-transitions are disabled. For both experimental configurations, we instantiate a new NFA for each of five random seeds and evaluate 100 randomly generated input strings, comparing the network’s binary activation mask at the final timestep to the exact set of reachable NFA states.

As summarized in Table~\ref{TABLE:SIMULATION-RESULTS}, across all seeds and both configurations, the binary state vector exactly matched the true reachable states with perfect accuracy. This empirically confirms Proposition~\ref{PROP:BINARY-STATE-VECTOR}, showing that the binary state vector in a TS-FFN corresponds precisely to the set of active NFA states—neither over- nor under-activating any state. Simultaneously, the results support Remark~\ref{REMARK:PARALLEL-PATH-TRACKING}: the TS-FFN implicitly tracks all parallel nondeterministic branches via matrix composition and binary thresholding, without simulating individual trajectories explicitly.

\subsection{Validating Theorem~\ref{THM:SUBSET-CONSTRUCTION}}
\label{EXP_6_3}

To validate Theorem~\ref{THM:SUBSET-CONSTRUCTION}, we compare the binary state vector trajectory of the TS-FFN at each timestep to the classical subset-construction trace of the corresponding NFA under nondeterministic transitions.  
$\varepsilon$-transitions are again disabled to focus on symbol-by-symbol evolution. For each of five random seeds under both configurations, we generate a new NFA and evaluate 100 randomly sampled input strings per configuration, comparing the complete trace of thresholded state vectors to the exact DFA-style reachable subsets computed by classical subset construction.

As shown in Table~\ref{TABLE:SIMULATION-RESULTS}, across all seeds and configurations, the TS-FFN’s thresholded trace matched the theoretical subset construction exactly. These results confirm that the network performs compositional updates that reproduce the deterministic subset evolution implied by Theorem~\ref{THM:SUBSET-CONSTRUCTION}, thereby providing empirical evidence for the constructive equivalence between symbolic subset composition and unrolled feedforward computation.

\subsection{Validating Lemma~\ref{LEMMA:EPSILON-CLOSURE}}
\label{EXP_6_4}

To validate Lemma~\ref{LEMMA:EPSILON-CLOSURE}, we test whether a TS-FFN correctly computes the $\varepsilon$-closure of an NFA state using repeated thresholded applications of the $\varepsilon$-transition matrix. For each of five random seeds under both configurations, we generate a new NFA with $\varepsilon$-transitions and select a random initial state. We then compare the closure vector computed by the network to the exact symbolic $\varepsilon$-closure set derived from the automaton’s transitive $\varepsilon$-reachability relation.

As shown in Table~\ref{TABLE:SIMULATION-RESULTS}, across all seeds and configurations, the TS-FFN’s computed closure matched the exact symbolic $\varepsilon$-closure with perfect accuracy. This empirically validates Lemma~\ref{LEMMA:EPSILON-CLOSURE}, demonstrating that the network’s fixed-point dynamics under repeated thresholded $\varepsilon$-matrix composition faithfully reproduce the full transitive closure over $\varepsilon$-arcs.

\subsection{Validating Theorem~\ref{THEOREM:FNN-NFA-SIMULATION}}
\label{EXP_6_5}

To validate Theorem~\ref{THEOREM:FNN-NFA-SIMULATION}, we evaluate whether a symbolic time-shared, depth-unrolled feedforward network (TS-FFN) can simulate the complete acceptance behavior of a given NFA, including both $\varepsilon$-transitions and nondeterministic branching. For each of five random seeds under both experimental configurations, we generate a distinct NFA with randomized transition structure, construct corresponding input/output training pairs, and initialize a TS-FFN whose transition matrices are deterministically set from the automaton’s symbolic transition definitions.

We then evaluate test-time acceptance predictions on 100 new strings per seed, comparing the TS-FFN’s output decisions with the exact acceptance results computed by the NFA. As summarized in Table~\ref{TABLE:SIMULATION-RESULTS}, the TS-FFN achieved near-perfect accuracy across both configurations. These findings empirically confirm that the symbolic TS-FFN, when initialized according to Theorem~\ref{THEOREM:FNN-NFA-SIMULATION}, performs exact simulation of NFA acceptance behavior, including full nondeterministic reachability and $\varepsilon$-closure computation.

\subsection{Validating Theorem~\ref{THEOREM:EQUIVALENCE-FNN-NFA}}
\label{EXP_6_6}

To empirically validate Theorem~\ref{THEOREM:EQUIVALENCE-FNN-NFA}, we test whether a symbolic time-shared, depth-unrolled feedforward network (TS-FFN) and its corresponding NFA accept exactly the same set of strings. For each trial, we randomly instantiate an NFA and construct its TS-FFN recognizer using the symbolic simulation framework introduced in Theorem~\ref{THEOREM:FNN-NFA-SIMULATION}. We then generate 100 random test strings per seed and compare the binary acceptance decision (accept/reject) of the TS-FFN with that of the original NFA.

As reported in Table~\ref{TABLE:SIMULATION-RESULTS}, all runs in the first configuration achieved perfect equivalence between the TS-FFN and the corresponding NFA. In the second configuration, the equivalence accuracy remained consistently high across all random seeds. These results confirm that the TS-FFN construction precisely preserves the automaton’s acceptance behavior, empirically validating Theorem~\ref{THEOREM:EQUIVALENCE-FNN-NFA} and establishing the TS-FFN as a complete and faithful simulator of nondeterministic finite automata.

\subsection{Validating Proposition~\ref{PROP:FNN-NFA-LEARNABILITY}}
\label{EXP:FNN-NFA-LEARNABILITY}

To empirically validate Proposition~\ref{PROP:FNN-NFA-LEARNABILITY}, we assess whether the acceptance behavior of an NFA can be \emph{learned} by training a time-shared, depth-unrolled feedforward network (TS-FFN) via gradient descent.  
While the theoretical formulation uses binary thresholding for exact symbolic interpretability, we also evaluate three alternative activation strategies—ReLU, sigmoid, and no activation (purely linear)—to examine practical optimization behavior. In all cases, the network follows the symbolic structure defined in Section~\ref{EXPERIMENTAL_SETUP}, treating the sparse transition matrices as trainable parameters.

Each network is randomly initialized using Kaiming initialization~\cite{he2015delvingdeeprectifierssurpassing} and trained for 30 epochs using the Adam optimizer~\cite{adammethodstochasticoptimization} with a learning rate of 0.001 and batch size of 6. Training minimizes binary cross-entropy loss between predicted and ground-truth acceptance labels, where the targets are obtained from the corresponding NFA.

Across five random seeds and both experimental configurations, we evaluate each trained model on 100 held-out test strings per seed. The results, summarized in Table~\ref{TABLE:EXP_THEOREM_5.1}, confirm that TS-FFNs are empirically learnable via gradient descent. Although binary thresholding successfully captures NFA behavior, the smoother activations (ReLU and sigmoid) yield improved convergence and slightly higher accuracy due to better gradient flow. The purely linear variant also performs strongly, demonstrating that even minimal nonlinear structure can recover automaton behavior through shared symbolic parameters. Together, these results support Proposition~\ref{PROP:FNN-NFA-LEARNABILITY}, distinguishing between \emph{constructive realizability} (guaranteed existence of exact parameters) and \emph{empirical learnability} (successful optimization in practice).

\begin{table}[h]
\centering
\caption{Test accuracy of TS-FFNs trained to replicate NFA acceptance behavior across activation strategies and configurations.}
\vspace{2mm}
\setlength{\tabcolsep}{3pt} 
\label{TABLE:EXP_THEOREM_5.1}
\begin{tabular}{ccccc}
\textbf{Activation} & \textbf{Config} & \textbf{Mean} & \textbf{Std. Dev.} & \textbf{95\% CI} \\
 & & \textbf{Accuracy} & & \\
\hline
Binary   & 1 & 0.9300 & 0.0908 & (0.8392,\ 1.0208) \\
No Activation & 1 & 0.9820 & 0.0403 & (0.9321,\ 1.0319) \\
ReLU & 1 & 0.9940 & 0.0134 & (0.9774,\ 1.0106) \\
Sigmoid & 1 & 0.9980 & 0.0045 & (0.9924,\ 1.0035) \\
\hline
Binary   & 2 & 0.8067 & 0.0709 & (0.7323,\ 0.8811) \\
No Activation & 2 & 0.8800 & 0.0583 & (0.8076,\ 0.9524) \\
ReLU     & 2 & 0.9320 & 0.0238 & (0.9024,\ 0.9616) \\
Sigmoid  & 2 & 0.9440 & 0.0231 & (0.9153,\ 0.9727) \\
\end{tabular}
\end{table}

\begin{figure}[t]
\centering
\includegraphics[width=0.70\textwidth]{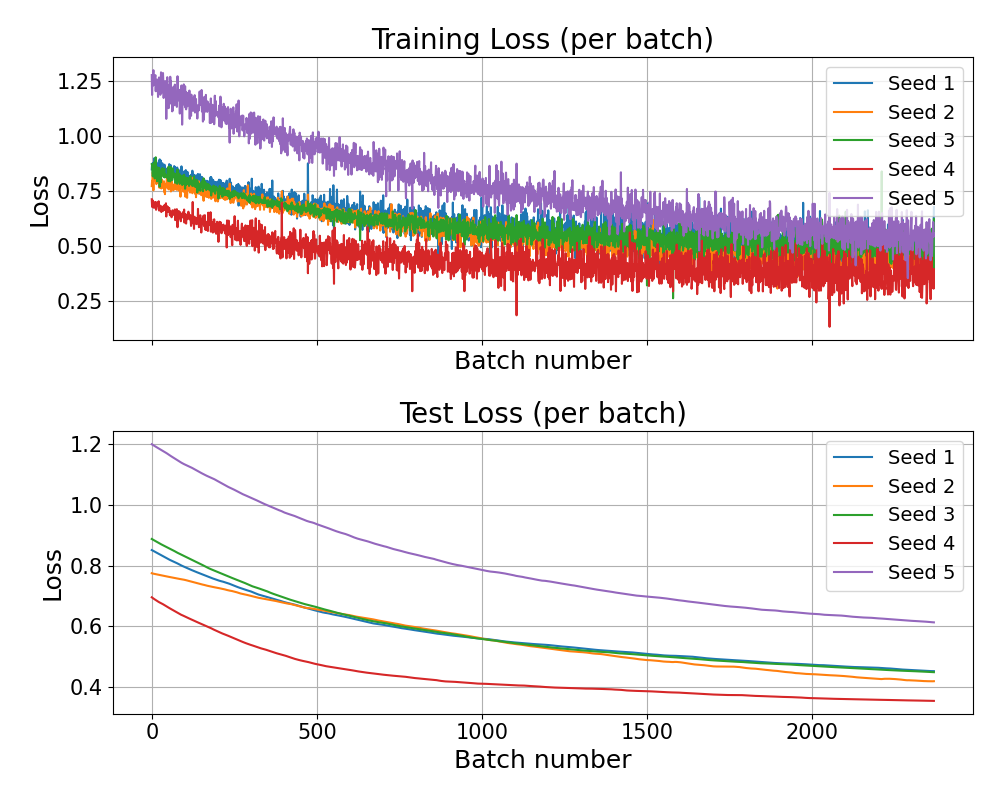}
\caption*{Figure 1: Batchwise training and test loss curves for \textbf{Configuration 1} with \textbf{Binary Activation} across all seeds.}
\label{fig:batchwise_nfa_losses_config_1_binary_activation}
\end{figure}

As shown in Fig.~\hyperref[fig:batchwise_nfa_losses_config_1_binary_activation]{1}-~\hyperref[fig:batchwise_nfa_losses_config_2_sigmoid]{8} provided in Appendix~\ref{figures}, both training and test losses decrease consistently across all seeds, configurations, and activation functions, confirming stable convergence and effective learning throughout the training process.

\section{Conclusion}
\label{sec:conclusion}

We have presented a constructive and formally exact framework for simulating nondeterministic finite automata (NFAs) using \emph{time-shared, depth-unrolled feedforward networks (TS-FFNs)}. The proposed formulation represents automaton transitions as sparse binary matrices, propagates nondeterministic branching through thresholded updates, and computes $\varepsilon$-closures via iterative matrix compositions—all within an acyclic, shared-parameter feedforward architecture. This construction establishes that every regular language can be simulated exactly by a TS-FFN with fixed parameter count independent of input length, offering a precise characterization of how neural architectures can realize symbolic computation.

Beyond theoretical realizability, we demonstrated \emph{empirical learnability}: TS-FFNs can be trained via standard gradient-based optimization to recover the acceptance behavior of NFAs from labeled examples. Across multiple configurations and activation strategies, these networks achieved high empirical agreement with ground-truth automata, validating the theoretical framework in practice. Together, these results unify automata theory and neural computation, showing that regular languages—traditionally confined to discrete symbolic models—can be both represented and learned within continuous, differentiable systems. This provides a principled foundation for neural–symbolic learning grounded in formal equivalence and constructive design rather than approximation or heuristic extraction.

By unifying two historically distinct paradigms—finite automata and neural networks—this work contributes toward a broader theoretical understanding of how modern deep architectures can embody and generalize symbolic structures.  
It marks a step toward a unified theory of neural computation that is interpretable, formally grounded, and compatible with classical models of computation.

\section{Limitations}
\label{sec:limitations}

While the proposed TS-FFN framework provides an exact and interpretable simulation of nondeterministic finite automata, its theoretical expressivity is currently restricted to the class of regular languages. Extending this constructive approach to more expressive language classes—such as context-free or context-sensitive languages—or to modern architectures such as transformers and recurrent state-space models remains an open direction for future research.

\section{Broader Impact}
\label{sec:broader-impact}

This work advances the theoretical understanding of how neural networks can implement symbolic computation, offering a rigorous bridge between automata theory and deep learning. The framework has potential implications for neural–symbolic reasoning, formal verification, program synthesis, natural language understanding, and interpretable AI systems. By enabling feedforward architectures to perform exact symbolic reasoning with transparent internal representations, this approach supports applications requiring correctness, verifiability, and alignment with formal logic. As a foundational theoretical contribution, the work is purely conceptual and does not pose foreseeable ethical or societal risks.

\bibliography{main}
\bibliographystyle{plainnat}

\appendix
\section{Appendix: Proofs of Theoretical Results}
\label{app:proofs}

\subsection{Proof Sketch of Proposition~\ref{PROP:BINARY-STATE-VECTOR}: Binary State Vector Representation}
\label{PROOF_PROP_1}
\begin{proof}[Proof Sketch]
Let $s_t \in \{0,1\}^n$ be the indicator vector of the current subset of active states at time $t$, and let $T^{x_t} \in \{0,1\}^{n \times n}$ be the binary transition matrix for input symbol $x_t$, where $T^{x_t}_{ij} = 1$ if and only if $q_j \in \delta(q_i, x_t)$.

We define the state transition update as:
\[
s_{t+1} = \mathbf{1}_{[T^{x_t} s_t > 0]}.
\]

We now verify that this update rule correctly computes the set of active states at time $t+1$.

Given $s_t$ as the binary indicator vector such that $[s_t]_j = 1$ if $q_j \in S_t$ (active), 0 otherwise, we have:
\[
[T^{x_t} s_t]_i = \sum_{j=1}^n T^{x_t}_{ij} [s_t]_j.
\]
This value is nonzero if and only if there exists some $j$ such that $q_j$ is active at time $t$ and $(q_j, x_t, q_i)$ is a valid transition—that is, $q_i \in \delta(q_j, x_t)$. Hence, $q_i$ is reachable from the current active states on symbol $x_t$.

Applying the elementwise thresholding $\mathbf{1}_{[\,\cdot\, > 0]}$ yields:
\[
[s_{t+1}]_i = \begin{cases}
1 & \text{if } q_i \in \delta(S_t, x_t), \\
0 & \text{otherwise}.
\end{cases}
\]
Thus, $s_{t+1}$ is the correct indicator vector for the new active state set after one input symbol.
\end{proof}

\subsection{Proof Sketch of Theorem~\ref{THM:SUBSET-CONSTRUCTION}: Subset Construction using Matrix-Based Thresholded Updates}
\label{PROOF_THEOREM_4_3}
\begin{proof}[Proof Sketch]
We proceed inductively.

\textbf{Base case.}  
At time $t = 0$, we initialize with $s_0 = e_{q_0}$, a one-hot vector corresponding to the singleton set $\{q_0\}$.

\textbf{Inductive step.}  
Suppose $s_{t-1}$ correctly encodes the subset of states $S_{t-1}$ reachable by consuming $x_1 \cdots x_{t-1}$. Let $T^{x_t}$ be the transition matrix for symbol $x_t$, with $T^{x_t}_{ij} = 1$ if $q_j \in \delta(q_i, x_t)$.

Then
\[
s_t = \mathbf{1}_{[T^{x_t} s_{t-1} > 0]}
\]
computes a new vector where $[s_t]_i = 1$ if and only if there exists $q_j \in S_{t-1}$ such that $q_i \in \delta(q_j, x_t)$.

From the definition of matrix-vector product and thresholding,
\[
[T^{x_t} s_{t-1}]_i = \sum_{j=1}^n T^{x_t}_{ij} [s_{t-1}]_j.
\]
The sum is nonzero if and only if some active state $q_j$ has a transition to $q_i$ on $x_t$. The thresholding function $\mathbf{1}_{[\cdot > 0]}$ converts this to a binary activation, thus encoding active states. Hence, $s_t$ is the correct indicator vector for $\delta(S_{t-1}, x_t) = S_t$.

By induction, $s_T$ represents $\delta(q_0, x_1 \cdots x_T)$, completing the proof.
\end{proof}

\subsection{Proof Sketch of Lemma~\ref{LEMMA:EPSILON-CLOSURE}: $\varepsilon$-Closure via Matrix-Based Thresholded Propagation}
\label{PROOF_LEMMA_2}
\begin{proof}[Proof Sketch]
Let $s^{(0)} \in \{0,1\}^n$ be the binary vector representing the initially active states at time $t$. Define
\[
s^{(k+1)} = \mathbf{1}_{[T^{\varepsilon} s^{(k)} > 0]}, \quad k \ge 0.
\]

\textbf{Step 1: Monotonicity.}  
Each application of $T^\varepsilon$ followed by thresholding adds new states to the active set (or leaves it unchanged), since
\[
(s^{(k+1)})_j = \mathbf{1}_{\left[\sum_{i=1}^n T^\varepsilon_{ji} (s^{(k)})_i > 0 \right]} \ge (s^{(k)})_j.
\]
Thus, the sequence $\{s^{(k)}\}$ is monotonic: $s^{(k)} \le s^{(k+1)}$ element-wise.

\textbf{Step 2: Finite Convergence.}  
Since each vector entry is binary and there are $n$ total states, the monotonic sequence $\{s^{(k)}\}$ can change at most $n$ times before reaching a fixed point.

\textbf{Step 3: Correctness.}  
At convergence, $s^{(K)}$ includes all states reachable from the original set via a chain of $\varepsilon$-transitions. Any such state is reachable via a finite path of length at most $n$, and will be activated through repeated matrix application. Conversely, no unreachable states will be activated because $T^\varepsilon$ is binary and has no negative entries or spurious transitions.

\textbf{Conclusion.}  
Therefore, $s^{(K)}$ represents the $\varepsilon$-closure of $s_t$, and $K \le n$.
\end{proof}

\subsection{Proof Sketch of Theorem~\ref{THEOREM:FNN-NFA-SIMULATION}: Simulation of NFAs via Time-Shared, Depth-Unrolled Feedforward Networks}
\label{PROOF_THEOREM_3}
\begin{proof}[Proof Sketch]
We are given an NFA $\mathcal{A} = (Q, \Sigma \cup \{\varepsilon\}, \delta, q_0, F)$ with $|Q| = n$ states. Let $s_t \in \{0,1\}^n$ denote the binary vector encoding which states are active after processing the first $t$ symbols of an input string $x = x_1 x_2 \dots x_L \in \Sigma^*$.

Let $s_{q_0} \in \{0,1\}^n$ be the one-hot vector with a 1 at the index corresponding to $q_0$ and 0 elsewhere.

Define:
\begin{itemize}
    \item $T^{x} \in \{0,1\}^{n \times n}$ such that $T^{x}_{ij} = 1$ iff $q_j \in \delta(q_i, x)$ (i.e., $q_i$ transitions to $q_j$ on symbol $x$);
    \item $T^{\varepsilon} \in \{0,1\}^{n \times n}$ such that $T^{\varepsilon}_{ij} = 1$ iff $q_j \in \delta(q_i, \varepsilon)$.
\end{itemize}

\textbf{Step 1: Initialization with $\varepsilon$-closure.}  
Compute the closure of the start state:
\[
s_0 = \mathbf{1}_{[(T^\varepsilon)^n s_{q_0} > 0]},
\]
where the closure is applied iteratively as established in Lemma~\ref{LEMMA:EPSILON-CLOSURE}.

\textbf{Step 2: Inductive simulation of transitions.}  
Suppose $s_{t-1}$ correctly represents the active state set after input prefix $x_1 \dots x_{t-1}$. Then we compute:
\[
\begin{aligned}
\tilde{s}_t &= \mathbf{1}_{[T^{x_t} s_{t-1} > 0]}, \quad \text{(symbol transition)}, \\
s_t &= \mathbf{1}_{[(T^\varepsilon)^n \tilde{s}_t > 0]}, \quad \text{($\varepsilon$-closure)}.
\end{aligned}
\]
By Lemma~\ref{LEMMA:EPSILON-CLOSURE}, $\tilde{s}_t$ encodes the direct successors under $x_t$, and $s_t$ then encodes all reachable states via $\varepsilon$-closure. This process exactly mirrors the operational semantics of an NFA.

Repeating this procedure for $t = 1, \dots, L$ yields the sequence $s_1, \dots, s_L$.

\textbf{Step 3: Final acceptance.}  
Let $\mathbf{1}_F \in \{0,1\}^n$ be the binary indicator vector of accepting states, with ${\mathbf{1}_F}_i = 1$ iff $q_i \in F$. The input string is accepted if and only if:
\[
\langle s_L, \mathbf{1}_F \rangle > 0,
\]
meaning that at least one accepting state becomes active by the end of processing the input. This completes the simulation.
\end{proof}

\subsection{Proof Sketch of Proposition~\ref{PROP:PARAMETER-EFFICIENCY}: Parameter Efficiency of TS-FFN Automata Simulators}
\label{PROOF_PROP_4_8}
\begin{proof}[Proof Sketch]
For each symbol $x \in \Sigma$, the symbolic transition is performed by a matrix–vector product followed by thresholding:
\[
s'_t = \mathbf{1}_{[T^x s_{t-1} > 0]}.
\]
To model $\varepsilon$-closure, we repeatedly apply thresholded matrix updates:
\[
\mathcal{E}(s) := \mathbf{1}_{[(T^\varepsilon)^n s > 0]},
\]
where $(T^\varepsilon)^n$ denotes at most $n$ iterations of the update rule $s^{(k+1)} = \mathbf{1}_{[T^\varepsilon s^{(k)} > 0]}$.

At each step, the network updates:
\[
s_t = \mathcal{E}\big(\mathbf{1}_{[T^{x_t} s_{t-1} > 0]}\big).
\]

The parameters involved are:
\begin{itemize}
    \item $k$ transition matrices $\{T^x\}_{x \in \Sigma}$, each of size $n \times n$,
    \item one $\varepsilon$-transition matrix $T^\varepsilon$,
    \item two vectors: $s_0 \in \mathbb{R}^n$ (initial state) and $f \in \mathbb{R}^n$ (accepting-state indicator).
\end{itemize}

Thus, the total number of symbolic or trainable parameters is at most $k n^2 + n^2 + 2n = \mathcal{O}(k n^2)$.  
Because these matrices are reused across all time steps and input length $L$, the parameter count remains fixed regardless of sequence length.
\end{proof}

\subsection{Proof Sketch of Theorem~\ref{THEOREM:EQUIVALENCE-FNN-NFA}: Equivalence between TS-FFNs and NFAs}
\label{PROOF_THEOREM_4_8}
\begin{proof}[Proof Sketch]

\textbf{Forward Direction.}  
Given a regular language \( \mathcal{L} \), let \( \mathcal{A} = (Q, \Sigma \cup \{\varepsilon\}, \delta, q_0, F) \) be an $\varepsilon$-NFA recognizing \( \mathcal{L} \) with $|Q| = n$.  
Construct a TS-FFN \( f_\theta \) as follows:
\begin{itemize}
    \item Represent the active subset of states at time \( t \) as a vector \( s_t \), initialized with a one-hot vector \( e_{q_0} \).
    \item Compute the $\varepsilon$-closure:
    \[
    s_0 = \mathbf{1}_{[(T^\varepsilon)^n e_{q_0} > 0]}.
    \]
    \item For each symbol \( x_t \) in the input string, compute:
    \[
    s_t = \mathbf{1}_{[(T^\varepsilon)^n \cdot \mathbf{1}_{[T^{x_t} s_{t-1} > 0]} > 0]}.
    \]
    \item Accept if \( \langle s_L, \mathbf{1}_F \rangle > 0 \), where \( \mathbf{1}_F \in \{0,1\}^n \) indicates the accepting states.
\end{itemize}

Because every operation corresponds exactly to the NFA’s transition function and $\varepsilon$-closure, the network accepts precisely those strings in \( \mathcal{L} \).

\textbf{Reverse Direction.}  
Given a TS-FFN \( f_\theta \) constructed by this symbolic procedure:
\begin{itemize}
    \item Each matrix \( T^x \in \{0,1\}^{n \times n} \) defines a transition relation: \( q_j \in \delta(q_i, x) \) iff \( T^x_{ji} = 1 \).
    \item Each $\varepsilon$-transition is defined analogously by \( T^\varepsilon_{ji} = 1 \iff q_j \in \delta(q_i, \varepsilon) \).
    \item The initial state corresponds to the one-hot vector \( e_{q_0} \), and the final states are encoded by \( \mathbf{1}_F \).
\end{itemize}

Therefore, the TS-FFN precisely simulates an NFA \( \mathcal{A}' \) whose accepted language satisfies
\[
f_\theta(x) = 1 \iff x \in \mathcal{L}(\mathcal{A}').
\]
\textbf{Conclusion.}  
The equivalence between TS-FFNs and NFAs holds constructively in both directions: any NFA can be simulated exactly by a TS-FFN, and any such TS-FFN corresponds to an NFA recognizing the same language.
\end{proof}

\subsection{Proof Sketch of Proposition~\ref{PROP:FNN-NFA-LEARNABILITY}: Realizability and Empirical Learnability of NFAs via TS-FFNs}
\label{PROOF_THEOREM_5_1}
\begin{proof}[Proof Sketch]
Let $\mathcal{A} = (Q, \Sigma \cup \{\varepsilon\}, \delta, q_0, F)$ be an NFA with $n = |Q|$ states, and let $D = \{(x_i, y_i)\}_{i=1}^m$ be a dataset labeled by its acceptance behavior.  
We construct a parameterized TS-FFN that follows the symbolic simulation described in Theorem~\ref{THEOREM:FNN-NFA-SIMULATION}, treating each transition matrix \( T^x \in \mathbb{R}^{n \times n} \) as a learnable parameter.

\textbf{Network definition.}
\begin{itemize}
  \item Let $e_{q_0} \in \mathbb{R}^n$ denote the one-hot vector corresponding to the start state $q_0$.
  \item Initialize with $\varepsilon$-closure:
  \[
  s_0 = \sigma\!\left((T^\varepsilon)^n e_{q_0}\right),
  \]
  where $\sigma$ is either a hard threshold $\mathbf{1}_{[z > 0]}$ or a differentiable proxy (e.g., sigmoid).  
  $(T^\varepsilon)^n$ denotes $n$ iterations ensuring closure convergence.
  \item For each symbol $x_t$ in $x = x_1 x_2 \dots x_L$:
  \[
  \tilde{s}_t = \sigma(T^{x_t} s_{t-1}), \quad
  s_t = \sigma\!\left((T^\varepsilon)^n \tilde{s}_t\right),
  \]
  with $s_t \in \mathbb{R}^n$ encoding active-state activations after $t$ steps.
  \item The final output is:
  \[
  f_\theta(x) = \mathrm{sigmoid}(\langle s_L, \mathbf{1}_F \rangle),
  \]
  where $\mathbf{1}_F \in \{0,1\}^n$ indicates the accepting states.
\end{itemize}

\textbf{Training objective.}
The network is trained with binary cross-entropy loss:
\[
\mathcal{L}(\theta) = -\sum_{i=1}^m \big[y_i \log f_\theta(x_i) + (1 - y_i) \log(1 - f_\theta(x_i))\big],
\]
optimized via gradient descent:
\[
\theta^{(k+1)} = \theta^{(k)} - \eta \nabla_\theta \mathcal{L}(\theta^{(k)}),
\]
where $\theta$ contains all entries of $\{T^x\}_{x \in \Sigma \cup \{\varepsilon\}}$.

\textbf{Discussion.}
\begin{enumerate}
  \item The TS-FFN is expressive enough to represent any NFA exactly (realizability).
  \item The loss $\mathcal{L}(\theta)$ provides a meaningful gradient signal whenever predictions differ from labels.
  \item Gradient descent empirically reduces $\mathcal{L}(\theta)$, yielding networks that approximate the target automaton’s behavior with high accuracy.
\end{enumerate}
We do not claim global convergence of gradient descent or the existence of a unique minimizer; such analysis is left for future work.  
Our claim is constructive existence plus empirical evidence of learnability observed in experiments.

\textbf{Conclusion.}
Therefore, a TS-FFN parameterization $\theta^\star$ exists that exactly simulates $\mathcal{A}$, and empirical optimization over data $D$ typically yields $\theta^*$ achieving high accuracy in replicating $\mathcal{A}$’s acceptance behavior.
\end{proof}

\section{Appendix: Example: Encoding Input Strings within the TS-FFN}
\label{EXAMPLE}
Consider an input string $x = aab$ and an automaton over $\Sigma = \{a, b\}$.  
The unrolled TS-FFN computes:
\[
\begin{aligned}
s_0 &= \mathbf{1}_{[(T^\varepsilon)^n \cdot e_{q_0} > 0]}, \\
s_1 &= \mathbf{1}_{[(T^\varepsilon)^n \cdot \mathbf{1}_{[T^a \cdot s_0 > 0]} > 0]}, \\
s_2 &= \mathbf{1}_{[(T^\varepsilon)^n \cdot \mathbf{1}_{[T^a \cdot s_1 > 0]} > 0]}, \\
s_3 &= \mathbf{1}_{[(T^\varepsilon)^n \cdot \mathbf{1}_{[T^b \cdot s_2 > 0]} > 0]}.
\end{aligned}
\]
Finally, acceptance is determined via a dot product with the indicator vector of accepting states:
\[
\text{Accept}(x) = \text{True} \iff \langle s_3, \mathbf{1}_F \rangle > 0.
\]
This explicit separation of control flow (symbol sequence) from computation is fundamental to the symbolic fidelity and modularity of the TS-FFN framework.  
It allows the model to retain both exact automata semantics and differentiable compositional structure.

\section{Appendix: Figures}
\label{figures}

\begin{figure}[t]
\centering
\includegraphics[width=0.70\textwidth]{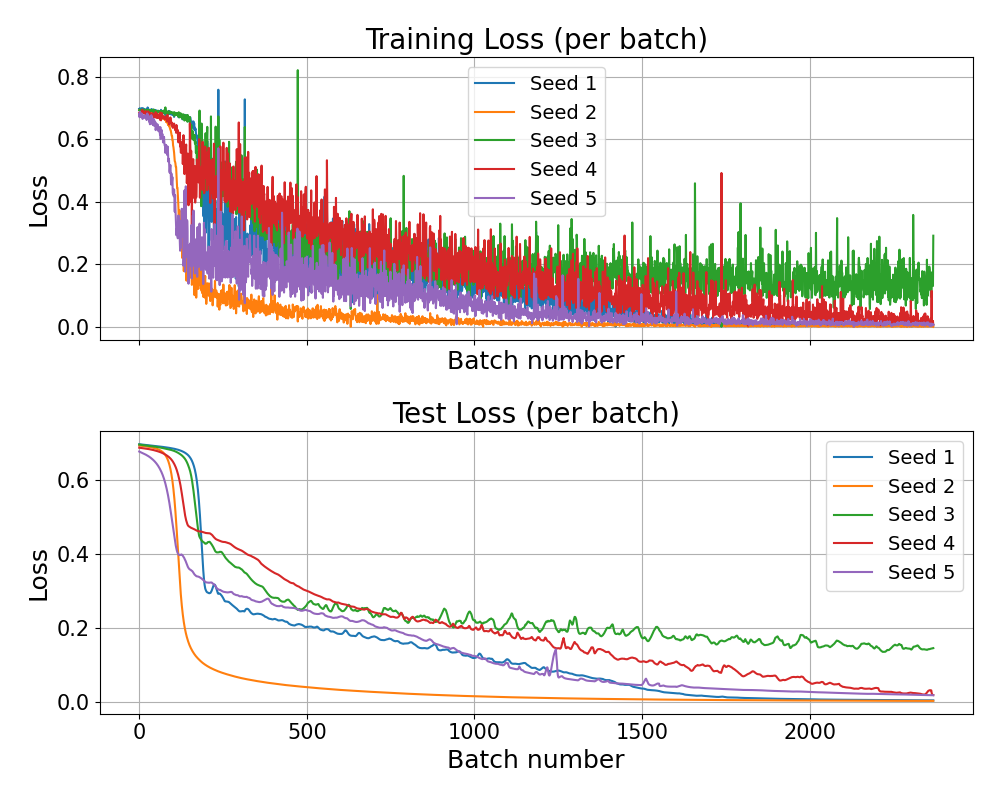}
\caption*{Figure 2: Batchwise training and test loss curves for \textbf{Configuration 1} with \textbf{No Activation (Linear)} across all seeds.}
\label{fig:batchwise_nfa_losses_config_1_no_activation}
\end{figure}

\begin{figure}[t]
\centering
\includegraphics[width=0.70\textwidth]{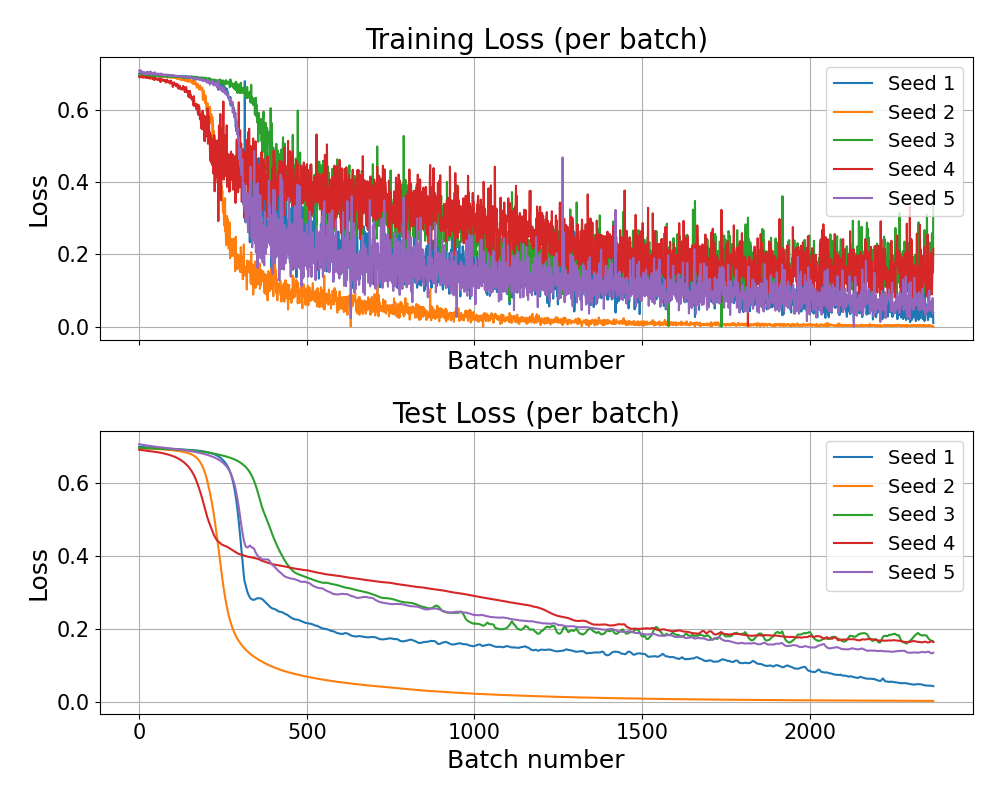}
\caption*{Figure 3: Batchwise training and test loss curves for \textbf{Configuration 1} with \textbf{ReLU Activation} across all seeds.}
\label{fig:batchwise_nfa_losses_config_1_relu}
\end{figure}

\begin{figure}[t]
\centering
\includegraphics[width=0.70\textwidth]{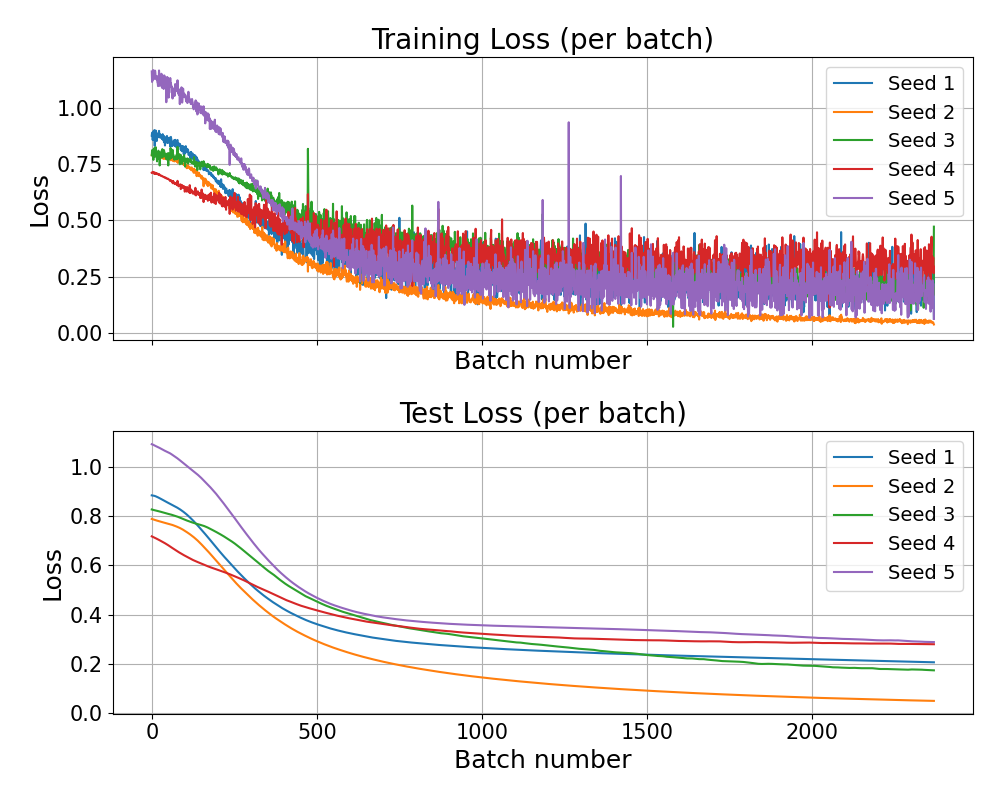}
\caption*{Figure 4: Batchwise training and test loss curves for \textbf{Configuration 1} with \textbf{Sigmoid Activation} across all seeds.}
\label{fig:batchwise_nfa_losses_config_1_sigmoid}
\end{figure}

\begin{figure}[t]
\centering
\includegraphics[width=0.70\textwidth]{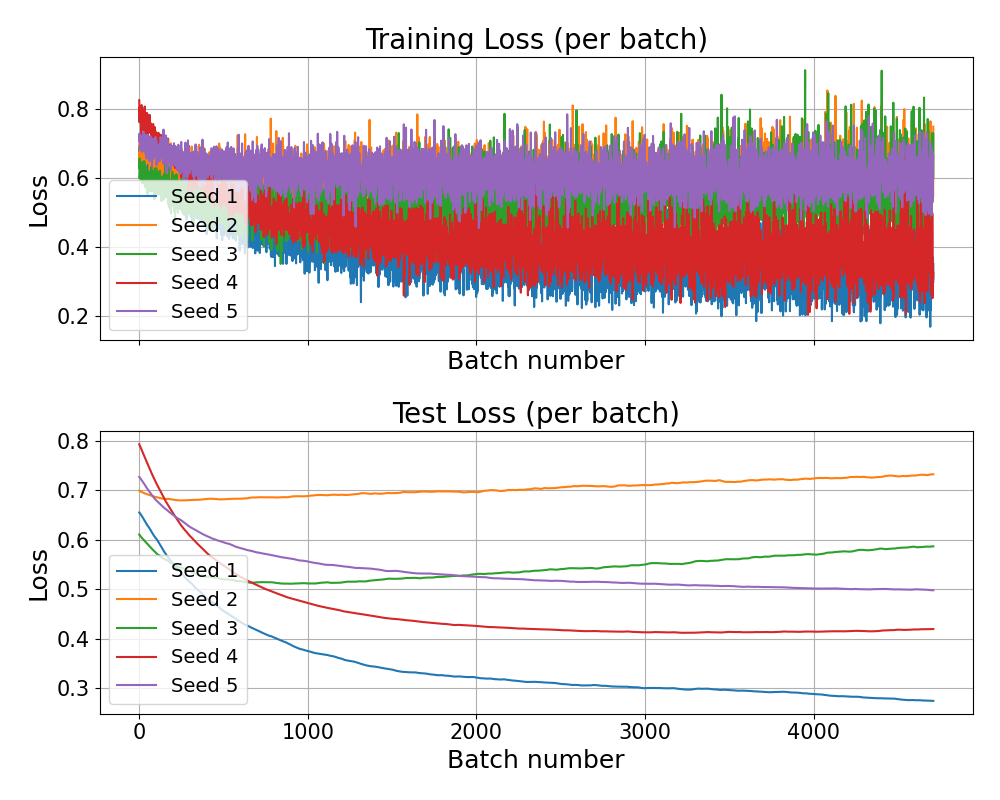}
\caption*{Figure 5: Batchwise training and test loss curves for \textbf{Configuration 2} with \textbf{Binary Activation} across all seeds.}
\label{fig:batchwise_nfa_losses_config_2_binary_activation}
\end{figure}

\begin{figure}[t]
\centering
\includegraphics[width=0.70\textwidth]{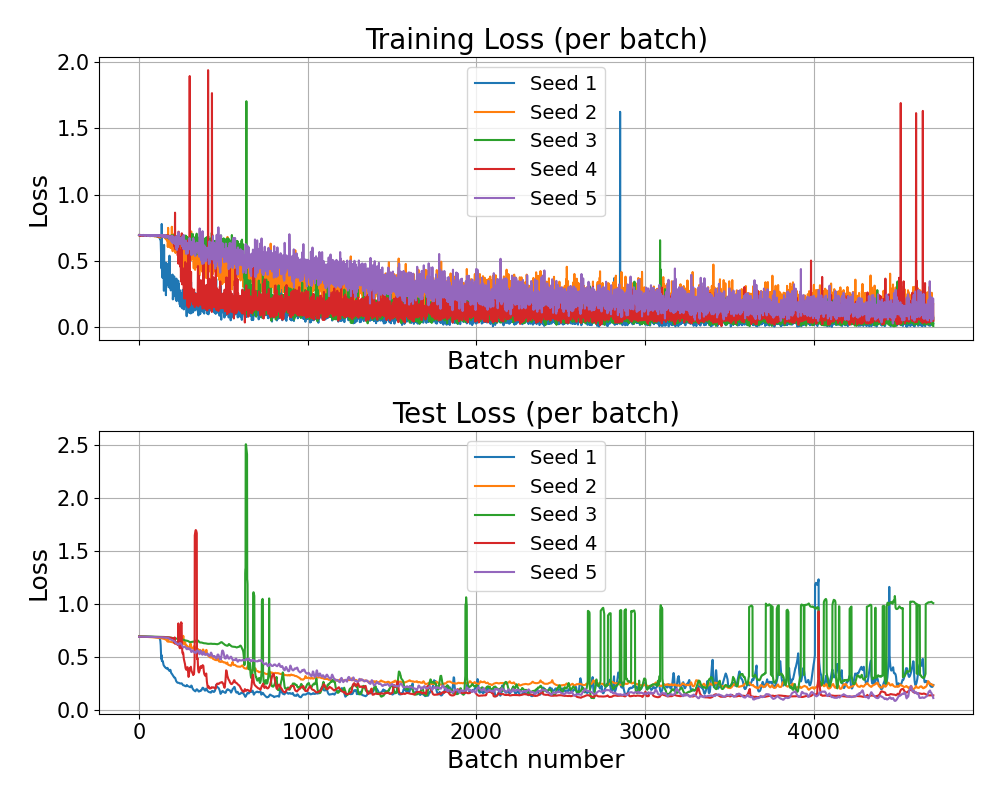}
\caption*{Figure 6: Batchwise training and test loss curves for \textbf{Configuration 2} with \textbf{No Activation (Linear)} across all seeds.}
\label{fig:batchwise_nfa_losses_config_2_no_activation}
\end{figure}

\begin{figure}[t]
\centering
\includegraphics[width=0.70\textwidth]{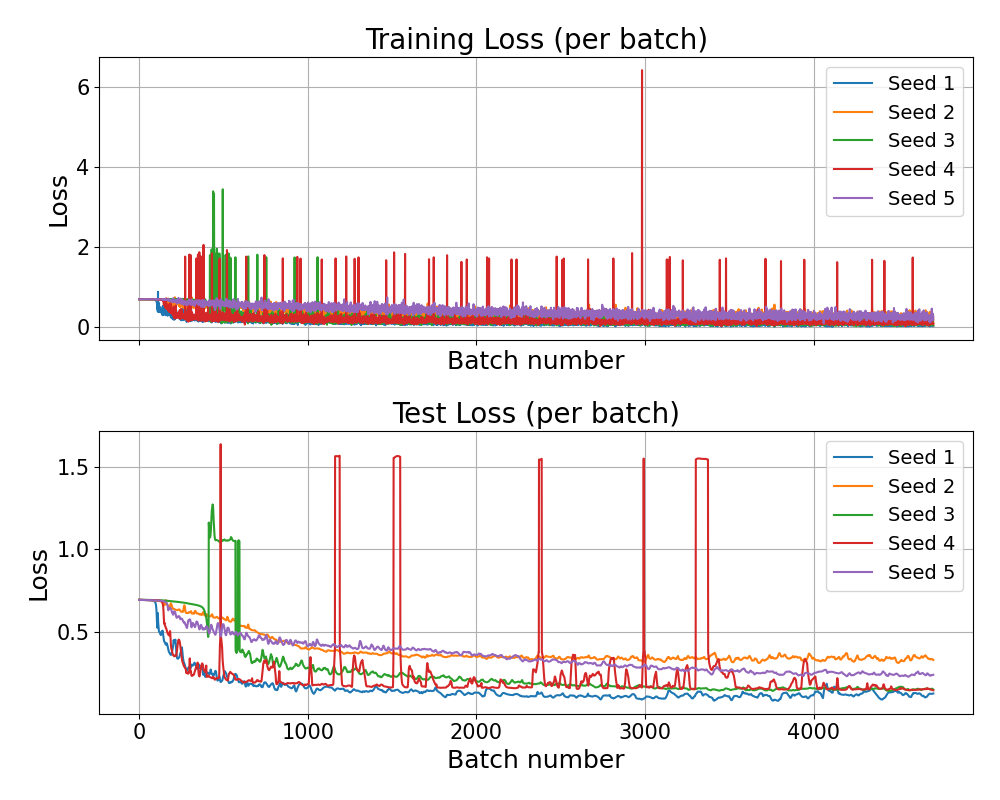}
\caption*{Figure 7: Batchwise training and test loss curves for \textbf{Configuration 2} with \textbf{ReLU Activation} across all seeds.}
\label{fig:batchwise_nfa_losses_config_2_relu}
\end{figure}

\begin{figure}[t]
\centering
\includegraphics[width=0.70\textwidth]{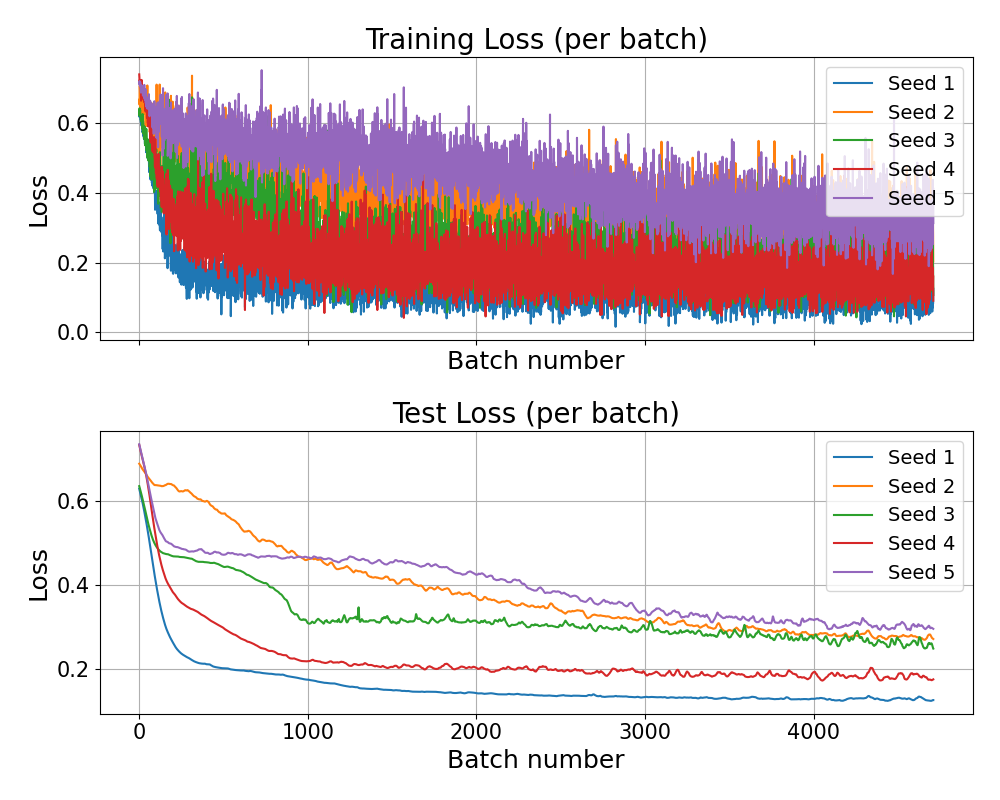}
\caption*{Figure 8: Batchwise training and test loss curves for \textbf{Configuration 2} with \textbf{Sigmoid Activation} acorss all seeds.}
\label{fig:batchwise_nfa_losses_config_2_sigmoid}
\end{figure}

\end{document}